\documentclass{article}
\pdfoutput=1

\usepackage[letterpaper]{geometry}
\usepackage{spconf,amsmath,epsfig}
\usepackage{hyperref}
\usepackage[caption=false,font=footnotesize]{subfig}
\usepackage{enumitem}
\usepackage[none]{hyphenat}

\begin{document}\sloppy

% Example definitions.
% --------------------
\def\x{{\mathbf x}}
\def\L{{\cal L}}

% Title.
% ------
\title{Why my photos look sideways or upside down? Detecting Canonical Orientation of Images using Convolutional Neural Networks}
%
% Single address.
% ---------------
\name{Kunal Swami, Pranav P Deshpande, Gaurav Khandelwal, Ajay Vijayvargiya}
\address{Samsung R\&D Insitute India - Bangalore \\
\{kunal.swami, pranav.ppd, gaurav.k7, ajay.v\}@samsung.com}
%
% For example:
% ------------
%\address{School\\
%	Department\\
%	Address\\
%   Email}
%
% Two addresses (uncomment and modify for two-address case).
% ----------------------------------------------------------
%\twoauthors
%  {A. Author-one, B. Author-two\sthanks{Thanks to XYZ agency for funding.}}
%	{School A-B\\
%	Department A-B\\
%	Address A-B}
%  {C. Author-three, D. Author-four\sthanks{The fourth author performed the work
%	while at ...}}
%	{School C-D\\
%	Department C-D\\
%	Address C-D\\
%   Email}
%

\maketitle

\begin{abstract}
Image orientation detection requires high-level scene understanding. Humans use object recognition and contextual scene information to correctly orient images. In literature, the problem of image orientation detection is mostly confronted by using low-level vision features, while some approaches incorporate few easily detectable semantic cues to gain minor improvements. The vast amount of semantic content in images makes orientation detection challenging, and therefore there is a large semantic gap between existing methods and human behavior. Also, existing methods in literature report highly discrepant detection rates, which is mainly due to large differences in datasets and limited variety of test images used for evaluation. In this work, for the first time, we leverage the power of deep learning and adapt pre-trained convolutional neural networks using largest training dataset to-date for the image orientation detection task. An extensive evaluation of our model on different public datasets shows that it remarkably generalizes to correctly orient a large set of unconstrained images; it also significantly outperforms the state-of-the-art and achieves accuracy very close to that of humans.
\footnote{\href{https://doi.org/10.1109/ICMEW.2017.8026216}{Link to official IEEE ICME 2017 Workshops paper}}
\end{abstract}
\begin{keywords}
Image orientation detection, deep learning, convolutional neural networks, transfer learning
\end{keywords}

\section{Introduction}
\label{sec:intro}
The proliferation of digital cameras has led to a significant increase in the number of photographs captured by people. While capturing images, the camera is not always held or mounted at correct angle, which results in image being displayed in wrong orientation. Modern digital cameras and smartphones have a built-in orientation sensor, which records the orientation of the camera during capture and writes it in the EXIF \cite{exif} data of the image. However, this technique is not consistently applied across different applications; many applications, such as default photo viewer in Windows $7$ doesn't support orientation tag. When an image is edited and saved using these applications, the orientation tag gets deleted, while in some cases the tag is not updated when image is rotated manually. Also, orientation sensor doesn't help when camera is aiming towards ground, for e.g., while capturing photos of documents and pictures kept on a table. The images captured by first person cameras, such as GoPro, which are mounted sideways and even upside-down, often require orientation correction. Automatic content creator software applications assume that input images are correctly oriented. Automatic detection and correction of image orientation is also useful in several image processing and computer vision systems. It is shown in \cite{jaderbergstn2015, chandrasekharcbir2016} that spatial transformations such as translations, scaling and especially rotations dwindle the accuracy of deep convolutional neural networks (CNNs). The traditional approach of making the systems transformation invariant doesn't work in case of large scale transformations. Therefore, an accurate image orientation detection and correction method is required to tackle aforementioned problems.

\subsection{Related Work}
\label{ssec:relatedwork}
Image orientation detection is a challenging task because digital images vary greatly in content (see Fig.~\ref{fig:sample4}, \ref{fig:sample6}, \ref{fig:sample13}, \ref{fig:sample14} and \ref{fig:sample15}). As a result, existing methods which mainly reckon on hand-engineered features for orientation detection are limited in their performance due to the intrinsic semantic gap between low-level vision features and high-level image semantics. Vailaya \textit{et al.} \cite{vailaya1999icip} first addressed the problem of image orientation detection using a Bayesian learning framework and spatial color moments as features. They reported an accuracy of $97\%$ on a high quality image set derived from Corel photos dataset. However, as stated in other works \cite{psychophysicalstudy2003,luollfsemanticintegration2004}, their remarkable accuracy was an artifact of the test dataset which mainly contained prototypical images. Later, Wang and Zhang \cite{wangzhang2001} using color moments and edge direction histogram features, obtained an accuracy of $78\%$ on another subset of Corel photos dataset. Zhang \textit{et al.} \cite{zhangindooroutdoor} treated indoor and outdoor images separately, Wang \textit{et al.} \cite{humanperceptioncues2003icip} integrated human perception cues, such as orientation of faces, position of sky, etc., into a Bayesian framework to obtain exact orientation angle of an image, reporting an accuracy of $94\%$ on a small test set of $1287$ images. Luo \textit{et al.} \cite{psychophysicalstudy2003} covered the psychophysical aspects of image orientation perception. Using the insights from \cite{psychophysicalstudy2003}, Luo and Boutell \cite{luollfsemanticintegration2004} integrated low-level features and several detectable semantic cues, such as faces, sky, grass, etc., into a Bayesian framework obtaining an accuracy of $90\%$ on a personal dataset of $3652$ images. The approach described in \cite{naturalimgstatistics} obtained accuracy close to $60\%$ on a personal test dataset. Baluja \cite{balujaboosting2007} used more than hundred classifiers trained with Adaboost and obtained maximum accuracy of $80.3\%$ on Corel photos dataset. Ciocca \textit{et al.} \cite{cioccallandfaces2010} incorporated faces as additional cue and obtained an accuracy of $86\%$ on a dataset of $4000$ online images. Cingovska \textit{et al.} \cite{hierarchicalcontent2011icip} used a hierarchical approach by first classifying images into their semantic group, such as faces, sky, etc. and then used a separately trained classifier for each semantic group to classify images into their correct orientation.

Evidently, highly discrepant detection rates have been reported in literature. The main reason for this discrepancy is large differences in test datasets; in some cases \cite{vailaya1999icip,humanperceptioncues2003icip,hierarchicalcontent2011icip} small and/or homogenous datasets are used for evaluation. Except \cite{cioccallandfaces2010}, all approaches have used rejection criteria with different rejection rates to achieve best accuracy. Due to these reasons, it is very difficult to ascertain the true performance of existing orientation detection methods. Furthermore, all existing methods have used highly imbalanced training and testing datasets in which  more than $50\%$ images were in correct ($0^\circ$) orientation; additionally \cite{luollfsemanticintegration2004,zhangindooroutdoor,naturalimgstatistics,balujaboosting2007,cioccallandfaces2010,hierarchicalcontent2011icip} completely ignored $180^\circ$ oriented images stating them as impractical. Psychophysical study \cite{psychophysicalstudy2003} says that humans are more likely to mis-orient images by $180^\circ$ orientation; therefore, removing this difficult case and using imbalanced datasets in which more than half images belong to correct orientation, considerably simplifies the orientation detection task. We argue that $180^\circ$ orientation is practically possible in case of first person cameras, such as GoPro as well as when camera is aiming towards ground.

Recently, Ciocca \textit{et al.} \cite{cioccalbp2015} used local binary patterns texture features with SVM as classifier and obtained $92\%$ accuracy for image orientation detection task. They addressed the problem of small and homogenous datasets by using SUN$397$ \cite{sun3972010} dataset for training and testing. Their method also outperformed other existing methods in literature by a significant margin. However, similar to other methods, they completely ignored $180^\circ$ orientation; the authors also used training and testing datasets that were highly imbalanced\textemdash more than $72\%$ images were in correct orientation. This made the orientation detection task considerably simpler. In this work, an extensive evaluation and comparison of Ciocca \textit{et al.}'s method reveals that their impressive accuracy was an artifact of their imbalanced training and testing datasets. It is also shown that their method doesn't generalize well to images outside SUN$397$ dataset. 
Apart from image orientation detection, there are other works in literature which focus on estimating (regressing) the exact skew angle of images \cite{salientline2013,orientationestimationcnn2015,fastorientationestimation2016, inplanerotationcnn2016}; however, it is a slightly different problem and their discussion is out of scope of this paper.

In this preliminary work, we use the CNN architecture proposed by Krizhevsky \textit{et al.} \cite{alexnet2012} (popularly known as AlexNet) and fine-tune it on the largest training dataset to-date for the image orientation detection task. Our extensive cross-dataset evaluation on several challenging scene and object recognition benchmark datasets \cite{holidays2008,mitindoor2009,pascalvoc2010} reveals that our model remarkably generalizes to correct orientation of a large variety of images with an impressive accuracy of $95\%$, which is very close to that of humans \cite{psychophysicalstudy2003}. Our model also significantly outperforms the current state-of-the-art method in \cite{cioccalbp2015} which reckons on hand-engineered features.

\subsection{Our Contributions}
\begin{itemize}
	\setlength\itemsep{0.0em}
	\item{As far as we know, this is the first work to leverage representational power of CNNs exclusively for image orientation detection task.}
	
	\item{We, unlike existing methods, do not ignore $180^\circ$ orientation and train as well as test our model on balanced datasets, therefore our model has no bias.}
	
	\item{We perform extensive evaluation of our model on challenging scene and object recognition benchmark datasets \cite{sun3972010,mitindoor2009,holidays2008,pascalvoc2010} to show its impressive generalizing capability. We didn't find such rigorous evaluation in any of the existing works in literature.}
	
	\item{Results show that our model significantly outperforms the current state-of-the-art method \cite{cioccalbp2015} and achieves $95\%$ accuracy, which is very close to that of humans \cite{psychophysicalstudy2003}.}
	
	\item{Lastly, we show visualizations of local image regions which are considered important by our model for classification \cite{selvarajugradcam2016}, helping us to compare its performance with human behavior.}	
\end{itemize}

\section{Datasets and Evaluation Protocol}
\label{sec:datasets}
In existing works, small and homogenous datasets have been used for training as well as evaluation. To address this, we derive our training set from the challenging scene recognition benchmark dataset SUN$397$ \cite{sun3972010} (similar to Ciocca \textit{et al.} \cite{cioccalbp2015}) and perform extensive cross-dataset evaluation using other challenging benchmark datasets in computer vision. SUN$397$ dataset has $397$ scene categories, each category having at least $100$ images and there are total $108,754$ images. For cross-dataset evaluation, we consider MIT Indoor \cite{mitindoor2009}, INRIA Holidays \cite{holidays2008} and Pascal VOC 2012 \cite{pascalvoc2010} datasets. We chose MIT Indoor dataset for testing because existing methods found difficulties with indoor images which contain lots of background clutter and lack discriminative features. INRIA Holidays dataset is a very good representative of real life images captured by people in their leisure time. Pascal VOC is an object-centric dataset compared to other datasets which contain only scene-centric images. As stated earlier, we use balanced testing datasets for evaluation, i.e., equal number of images for each orientation category to eliminate the effect of any bias. Additionally, we compare the performance of our model with current state-of-the-art method of Ciocca \textit{et al.} \cite{cioccalbp2015} on aforementioned datasets.

\section{Proposed Method}
\label{sec:proposedmethod} 
Our main focus in this work is to bridge the semantic gap between existing image orientation detection methods and human behavior. Astonished by the recent success of CNNs in challenging computer vision tasks \cite{alexnet2012,objemerge2014,midlevelrepresentations}, we decided to leverage their representational power for image orientation detection. For this task, it is possible to create a large training dataset and train the network from scratch; however, it is well known that pre-training a CNN on a large corpus of \emph{outside data} and fine-tuning it on the \emph{target data} not only helps the model to converge faster, but also results in significant performance boost \cite{midlevelrepresentations,transferablefeatures2014}, assuming that the outside and target data are of similar visual characteristics. Therefore, to this end, we decided to choose AlexNet CNN model proposed by Krizhevsky \textit{et al.} \cite{alexnet2012} and pre-trained on the MIT Places dataset \cite{places22016}, specifically, Places$365$ dataset which comprises of $1.8$ million images from $365$ scene categories. ImageNet \cite{imagenet2015} dataset has object-centric images which are quite dissimilar to our training dataset; moreover, Zhou \textit{et al.} \cite{objemerge2014} discovered that CNNs trained to perform scene classification implicitly learned to detect objects. Therefore, we found CNN pre-trained on Places365 dataset as a better choice for our task. 

We restrict image rotation to following four angles: $0^{\circ}$, $90^{\circ}$, $180^{\circ}$ and $270^{\circ}$. We argue that this kind of coarse orientation correction suffices for majority of user-centric use-cases; moreover, orientation correction at finer-level has deteriorating effects of image cropping. However, when it is required to determine the exact \textit{skew} angle of an image, we favor a hierarchical approach in which an image is first correctly oriented to one of the four aforementioned orientation angles. This approach minimizes the search for correct skew angle to a reasonable range (say, $\pm45^{\circ}$), thus, preventing erroneous estimations.  This approach is also consistent with the assumption of current skew detection algorithms in \cite{salientline2013,fastorientationestimation2016}.

\subsection{CNN Architecture}
\label{ssec:cnnarchitecture}
Our CNN architecture is inspired by AlexNet \cite{alexnet2012} and pre-trained on Places$365$ dataset \cite{places22016}. The network has five convolution (\textit{conv}) layers which are activated by rectfied linear units (ReLU), max-pooling is applied after $1^{st}$, $2^{nd}$ and $5^{th}$ convolution layers. Local response normalization is applied after $1^{st}$ and $2^{nd}$ convolution layers. Layers $6$, $7$ are fully connected (\textit{fc}) layers and layer $8$ is a softmax layer. Our network accepts $256$x$256$x$3$ size images as input. The last output layer \textit{fc8} was removed and replaced by the one with four outputs for our task. Similar to \cite{alexnet2012}, dropout with rate $0.5$ was implemented after \textit{fc6} and \textit{fc7} to control overfitting. The remaining parameters of the network remained same as in \cite{alexnet2012}.

\begin{figure*}[t]
	\centering
	
	\subfloat{\includegraphics[width=0.09\textwidth]{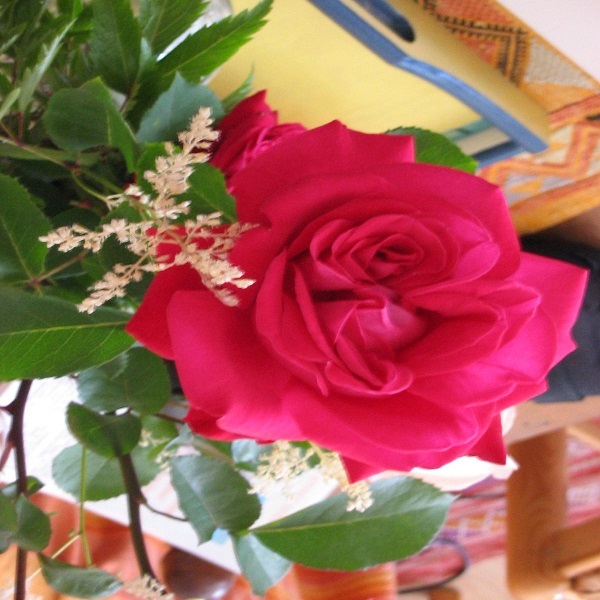}
		\label{fig:sample1}}
	\subfloat{\includegraphics[width=0.09\textwidth]{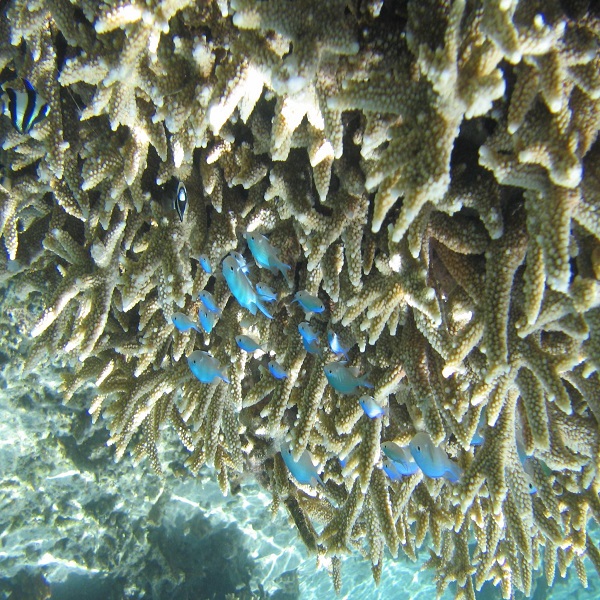}
		\label{fig:sample2}}	
	\subfloat{\includegraphics[width=0.09\textwidth]{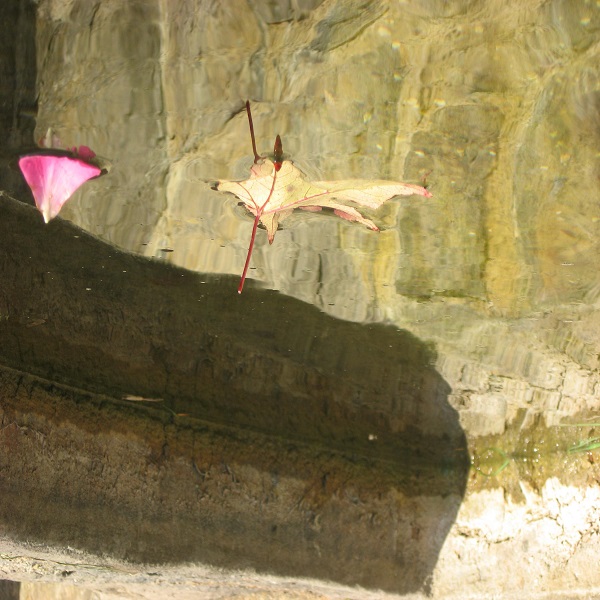}
		\label{fig:sample4}}
	\subfloat{\includegraphics[width=0.09\textwidth]{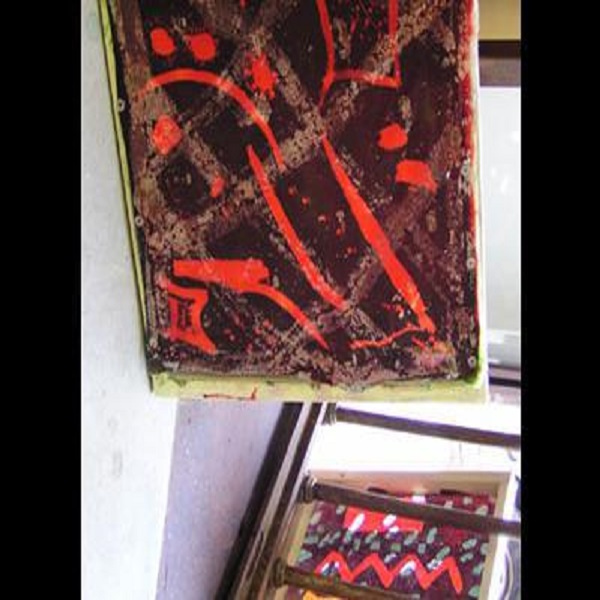}
		\label{fig:sample6}}
	\subfloat{\includegraphics[width=0.09\textwidth]{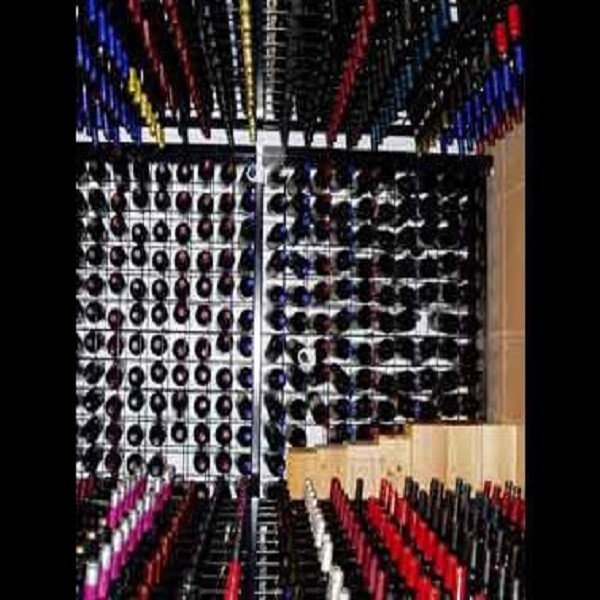}
		\label{fig:sample7}}
	\subfloat{\includegraphics[width=0.09\textwidth]{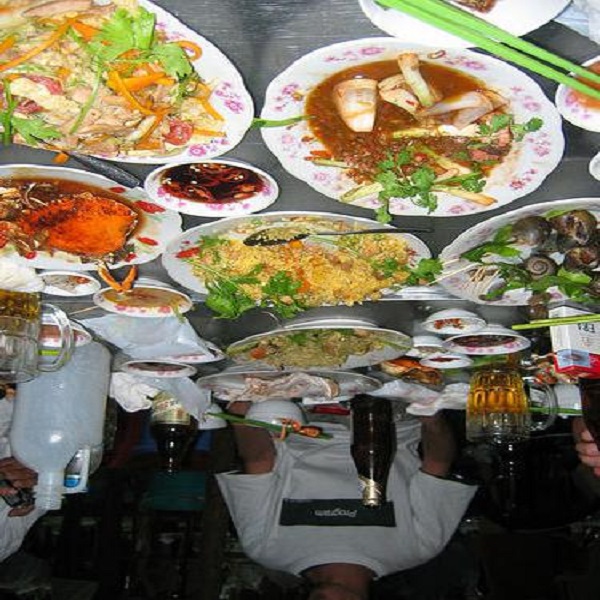}
		\label{fig:sample8}}
	\subfloat{\includegraphics[width=0.09\textwidth]{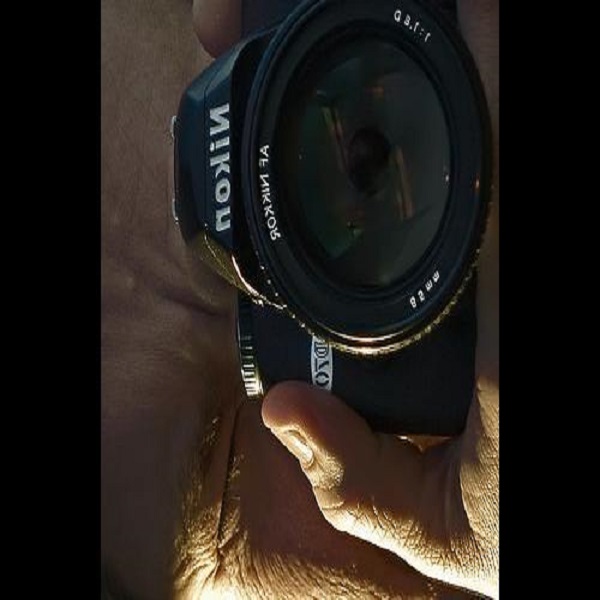}
		\label{fig:sample9}}
	\subfloat{\includegraphics[width=0.09\textwidth]{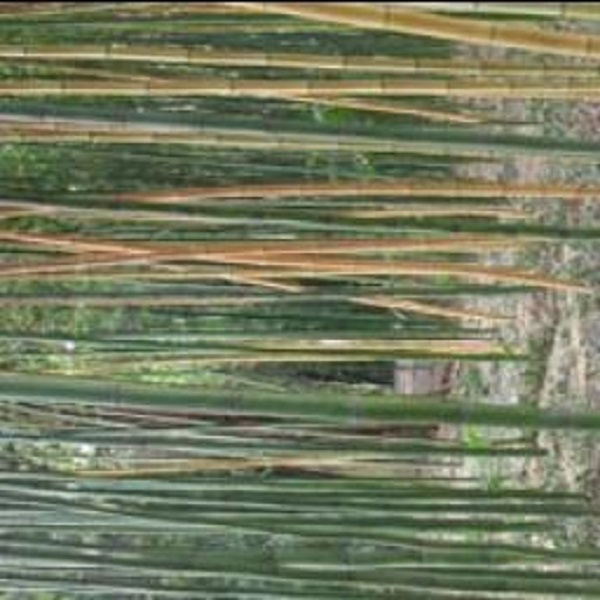}
		\label{fig:sample13}}
	\subfloat{\includegraphics[width=0.09\textwidth]{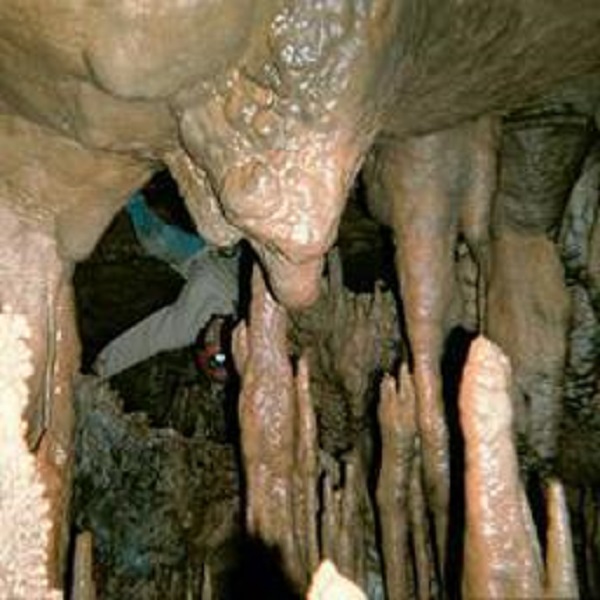}
		\label{fig:sample14}}
	\subfloat{\includegraphics[width=0.09\textwidth]{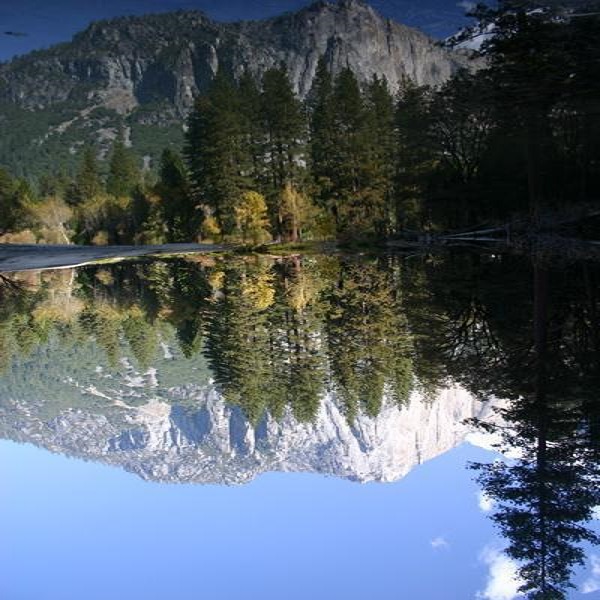}
		\label{fig:sample15}}

	\subfloat{\includegraphics[width=0.09\textwidth]{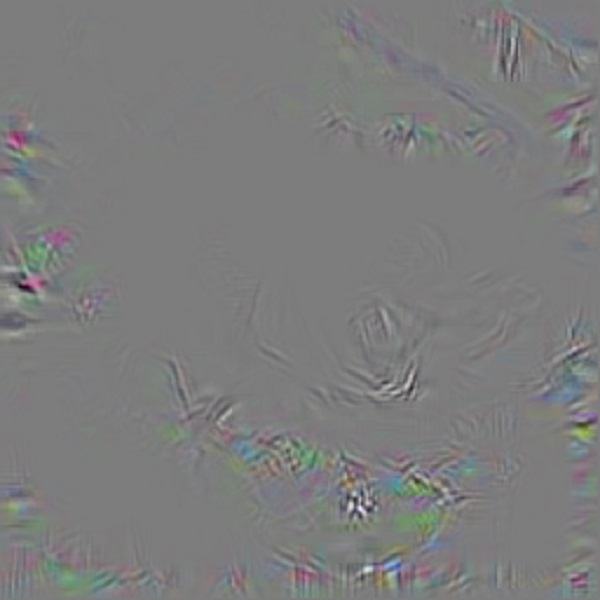}
		\label{fig:samplengm1}}
	\subfloat{\includegraphics[width=0.09\textwidth]{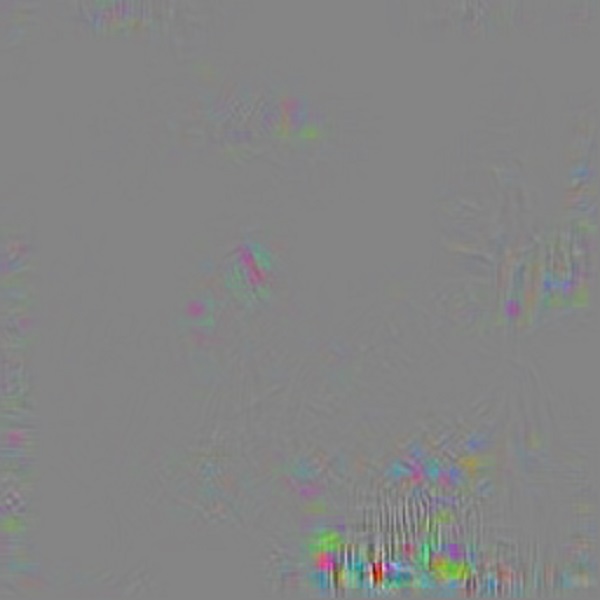}
		\label{fig:samplengm2}}
	\subfloat{\includegraphics[width=0.09\textwidth]{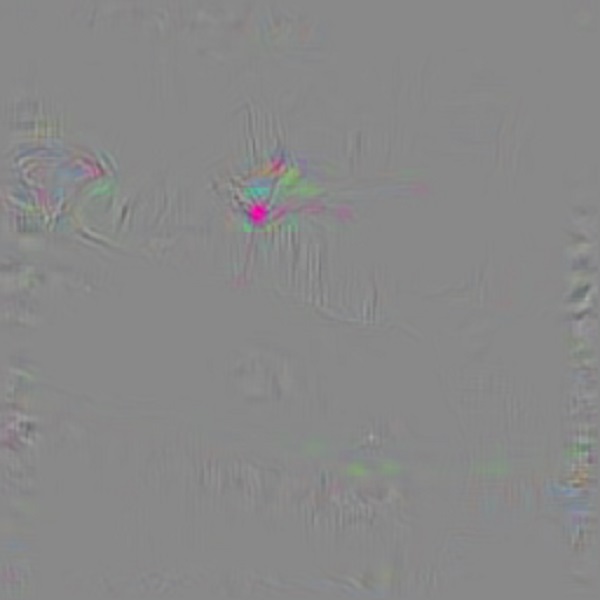}
		\label{fig:samplengm4}}
	\subfloat{\includegraphics[width=0.09\textwidth]{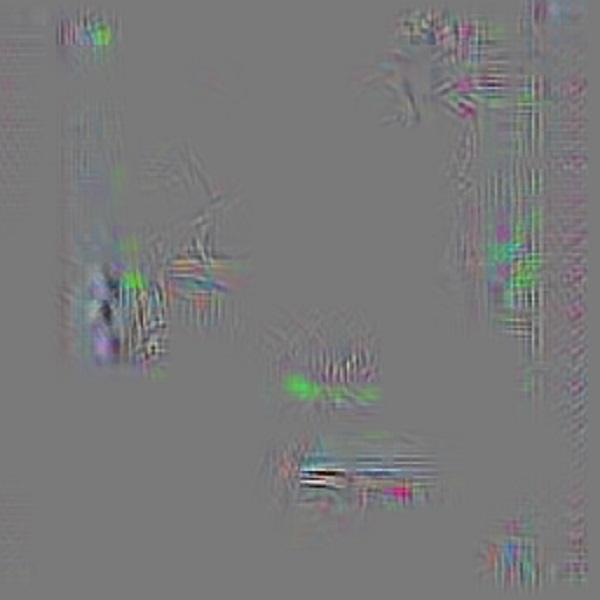}
		\label{fig:samplengm6}}
	\subfloat{\includegraphics[width=0.09\textwidth]{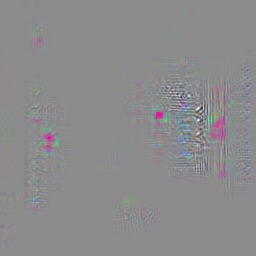}
		\label{fig:samplengm7}}
	\subfloat{\includegraphics[width=0.09\textwidth]{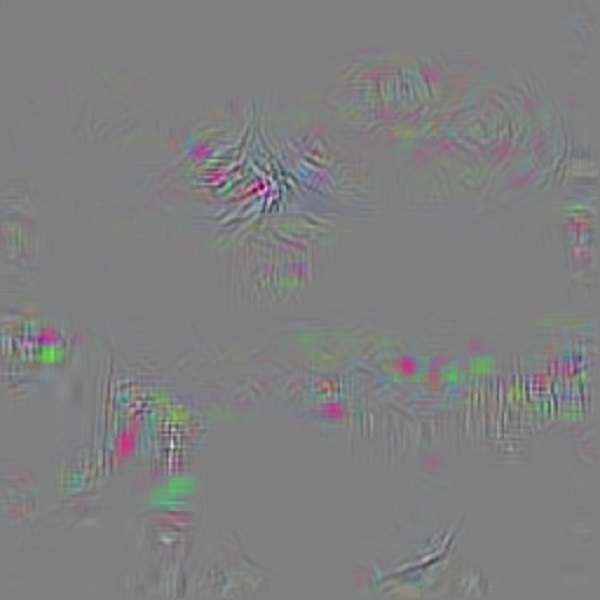}
		\label{fig:samplengm8}}
	\subfloat{\includegraphics[width=0.09\textwidth]{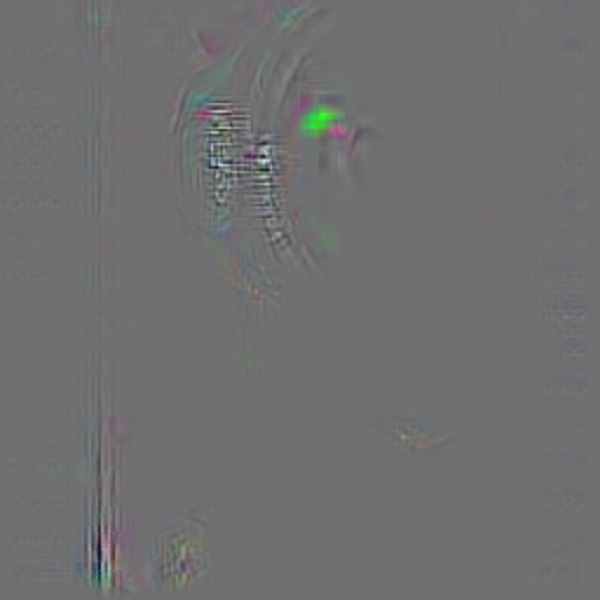}
		\label{fig:samplengm9}}
	\subfloat{\includegraphics[width=0.09\textwidth]{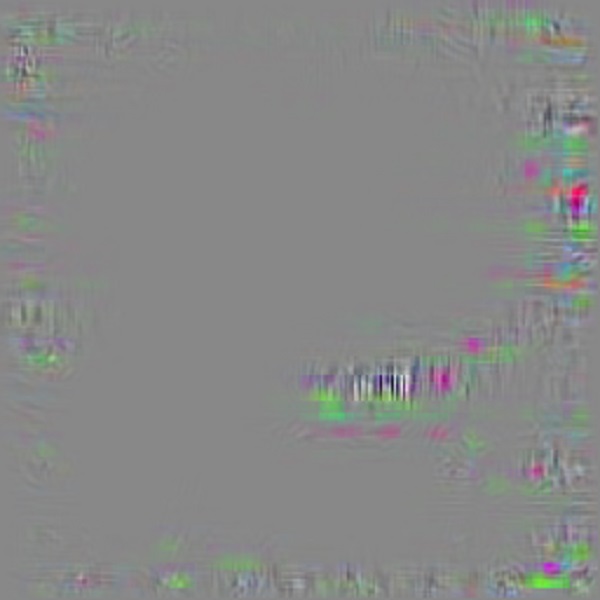}
		\label{fig:samplengm13}}
	\subfloat{\includegraphics[width=0.09\textwidth]{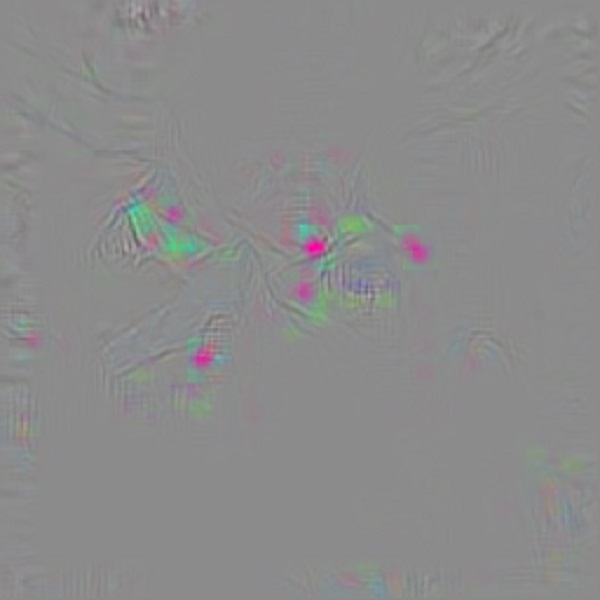}
		\label{fig:samplengm14}}
	\subfloat{\includegraphics[width=0.09\textwidth]{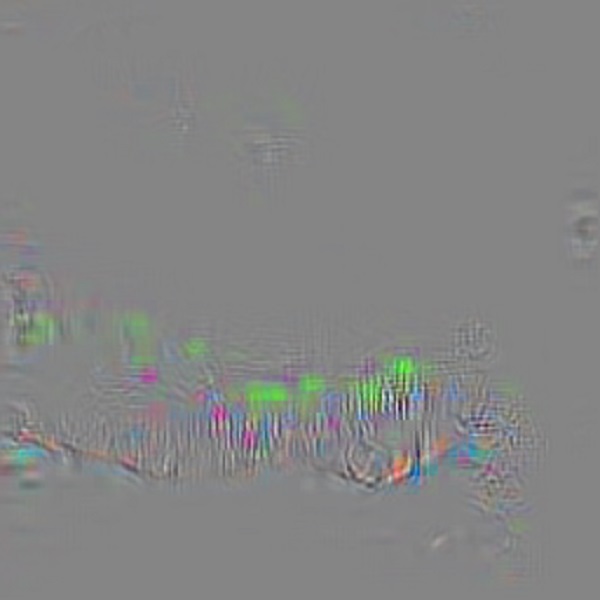}
		\label{fig:samplengm15}}

	\setcounter{subfigure}{0}
	\subfloat[$180^{\circ}$]{\includegraphics[width=0.09\textwidth]{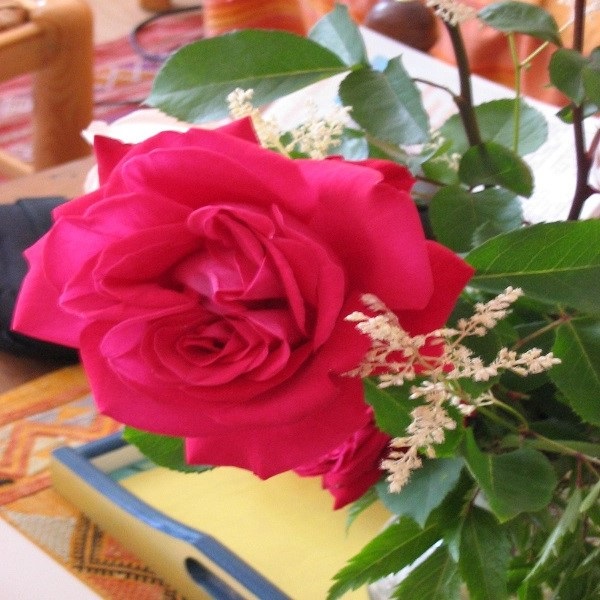}
		\label{fig:samplecorrect1}}
	\subfloat[$180^{\circ}$]{\includegraphics[width=0.09\textwidth]{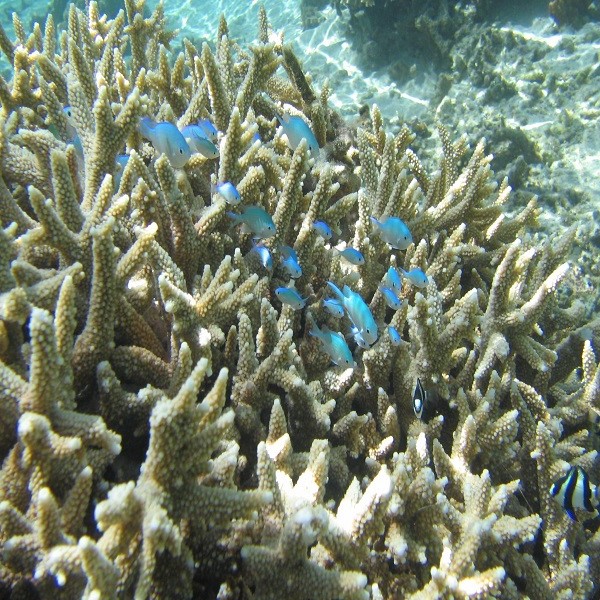}
		\label{fig:samplecorrect2}}
	\subfloat[$180^{\circ}$]{\includegraphics[width=0.09\textwidth]{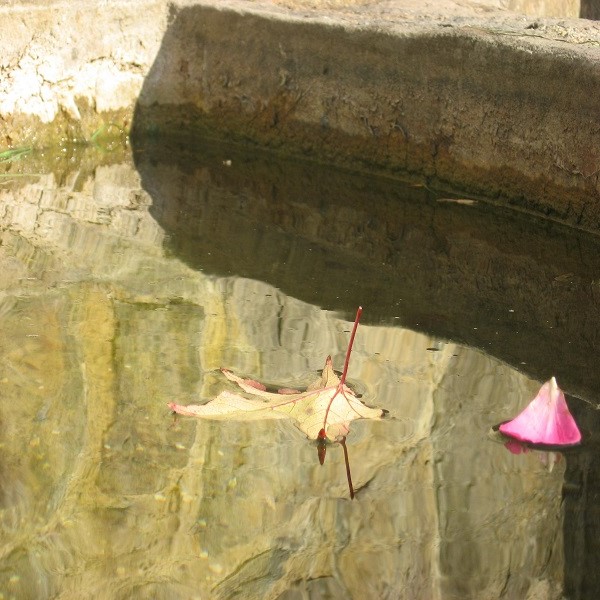}
		\label{fig:samplecorrect4}}
	\subfloat[$90^{\circ}$]{\includegraphics[width=0.09\textwidth]{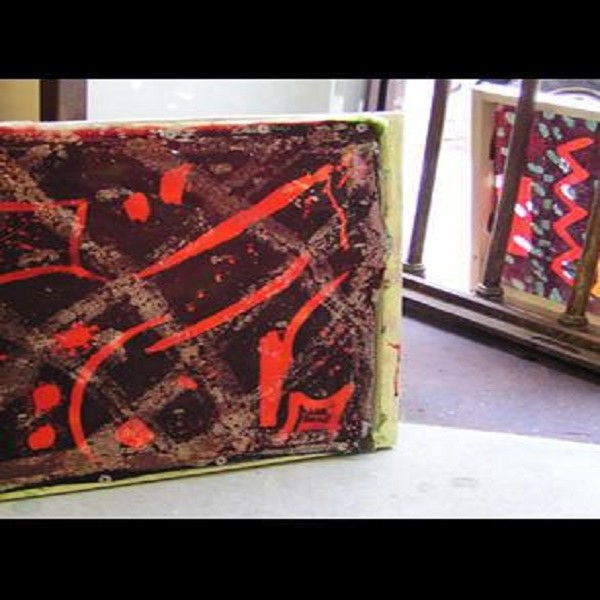}
		\label{fig:samplecorrect6}}
	\subfloat[$270^{\circ}$]{\includegraphics[width=0.09\textwidth]{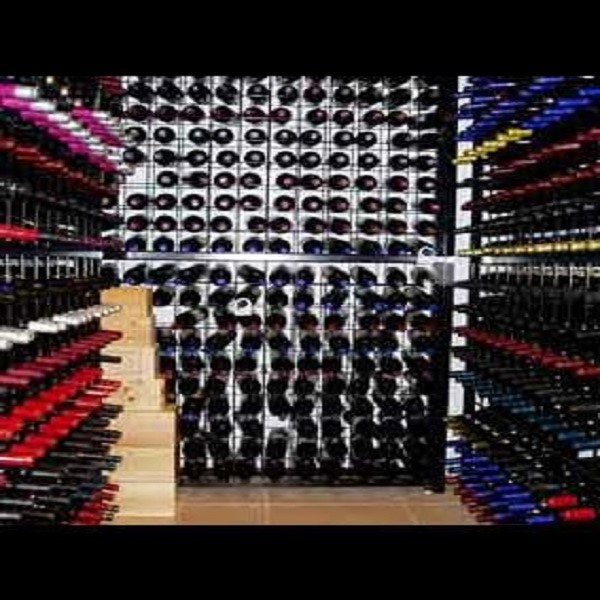}
		\label{fig:samplecorrect7}}
	\subfloat[$180^{\circ}$]{\includegraphics[width=0.09\textwidth]{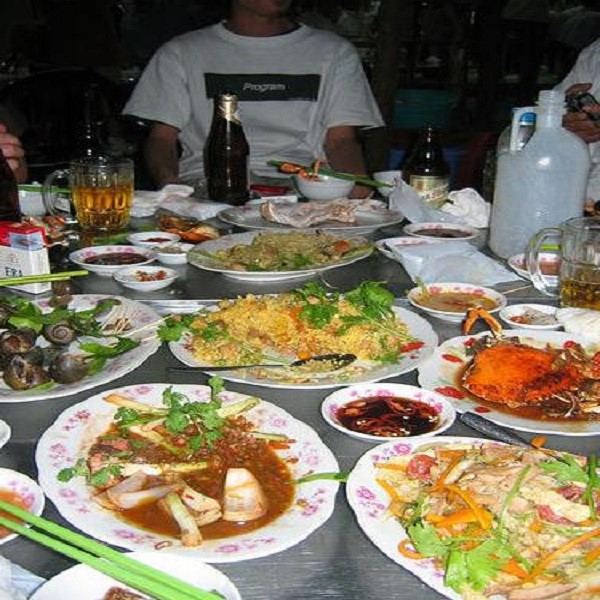}
		\label{fig:samplecorrect8}}
	\subfloat[$270^{\circ}$]{\includegraphics[width=0.09\textwidth]{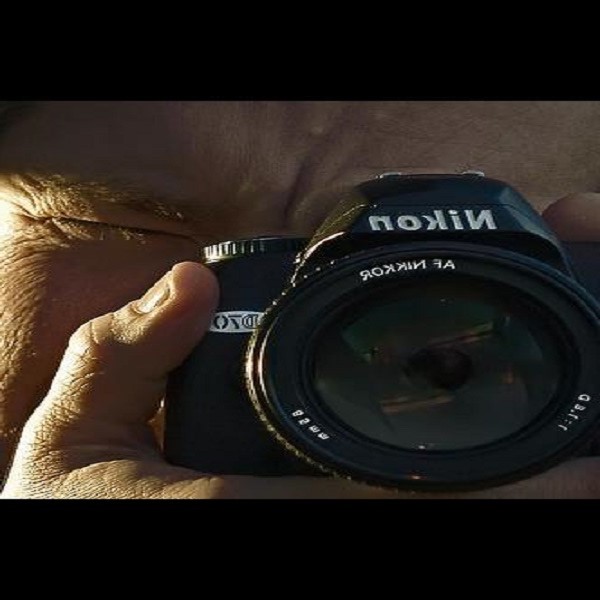}
		\label{fig:samplecorrect9}}
	\subfloat[$270^{\circ}$]{\includegraphics[width=0.09\textwidth]{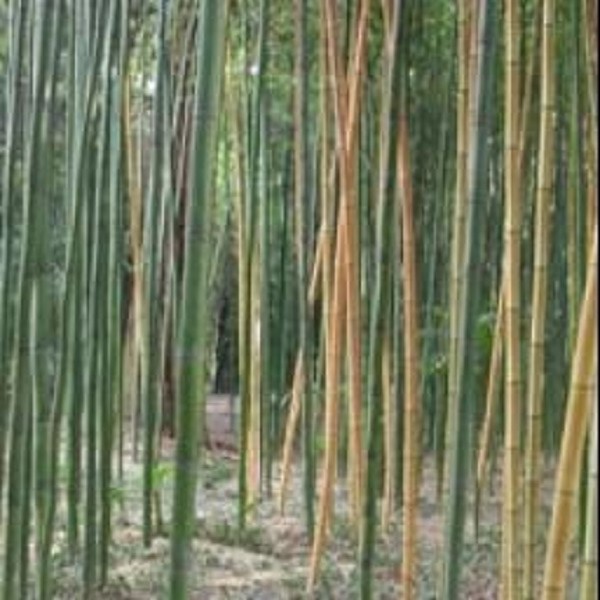}
		\label{fig:samplecorrect13}}
	\subfloat[$180^{\circ}$]{\includegraphics[width=0.09\textwidth]{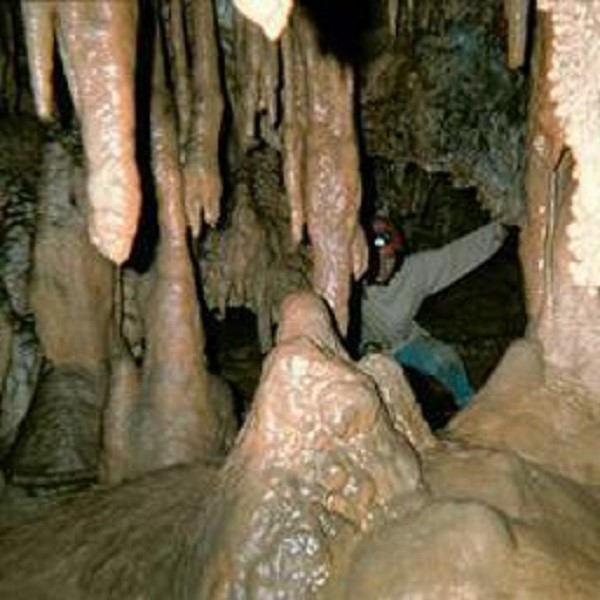}
		\label{fig:samplecorrect14}}
	\subfloat[$180^{\circ}$]{\includegraphics[width=0.09\textwidth]{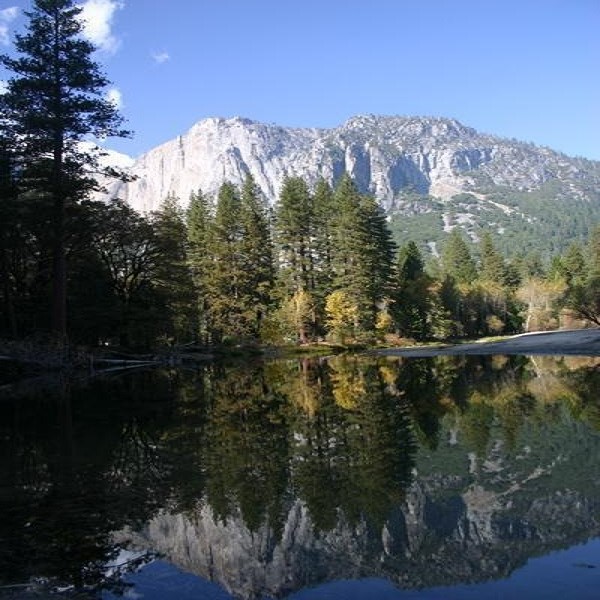}
		\label{fig:samplecorrect15}}

	\caption{Qualitative results of our method. First row shows rotated input images. Second row images show discriminative regions from corresponding first row images identified by our model for orientation classification task. Third row shows images rotated according to predicted orientation label which is written in the caption. Images best viewed in color.}
	\label{fig:qualitativeresults}
\end{figure*}

\subsection{Training}
\label{ssec:training}
We used $45,000$ images from SUN$397$ dataset for training, all images were initially in their correct orientation. For training, we additionally rotate each training image by $90^{\circ}$, $180^{\circ}$ and $270^{\circ}$ degrees and label them accordingly. Let $\mathcal{D} = \{(\mathcal{I}_i, \theta) \text{ } | \text{ } i\in[1,N] \text{, } \theta \in \{0,1,2,3\}\}$ be the training dataset. Here, $N$ is the number of training samples, which is $180,000$ in our case. The class label $\theta$ denotes the correct orientation of an input image\textemdash $0$ for $0^{\circ}$, $1$ for $90^{\circ}$, $2$ for $180^{\circ}$ and $3$ for $270^{\circ}$. Let $z$ be the four dimensional vector representing final softmax layer (\textit{fc8}) of the network, with $z_j$ denoting output at $j^{th}$ unit. Therefore, probability that the class label of $\mathcal{I}_i$ training sample is $j$, is calculated as: 

\begin{equation}
P(\theta=j\text{ }|\text{ }\mathcal{I}_i) = \frac{e^{z_j}}{\sum_{k=1}^{4} e^{z_k}} 
\end{equation}

The corresponding simplified cross-entropy loss ($L_i$) for this $i^{th}$ training sample is given by:

\begin{equation}
L_i = -\log(\frac{e^{z_j}}{\sum_{k=1}^{4} e^{z_k}})
\end{equation}

The learning task of our four-class classification problem is to minimize the above cross-entropy loss over the entire training dataset. Since \textit{conv1}, \textit{conv2} and \textit{conv3} contain generic low-level features, such as Gabor filters and color blobs \cite{transferablefeatures2014,visualizecnn2014}, we kept these layers intact. First, we tried fine-tuning only the fully-connected layers \textit{fc6} and \textit{fc7}. It is shown in \cite{transferablefeatures2014,midlevelrepresentations} that higher layers (\textit{fc6}, \textit{fc7} and \textit{fc8}) learn features which are task-specific and are non-transferable, while features learned by middle-level convolution layers (\textit{conv4}, \textit{conv5}) are transferable and can be fine tuned. Out of all different experiments, fine-tuning \textit{conv4}, \textit{conv5} layers and training \textit{fc6}, and \textit{fc7} layers from scratch performed best for us. In one experiment, we removed \textit{fc7} layer and reduced dimension of \textit{fc6} to $1024$; however, it lead to slight decrease ($\approx1.5\%$) in overall accuracy. We trained our model using the stochastic gradient descent (SGD) method with momentum $0.9$ and batch size of $256$. The non-fine-tuning layers (\textit{fc6}, \textit{fc7}, \textit{fc8}) were initialized by zero-mean Gaussian distribution with $0.01$ standard deviation. We initialized learning rate of fine tuning (\textit{conv4}, \textit{conv5}) as well as non-finetuning layers (\textit{fc6}, \textit{fc7}, \textit{fc8}) to $0.01$ and used overall network learning rate as $5$x$10^{-4}$ (see \cite{caffe2014}) with weight decay $0.0005$. This was to prevent significant weight changes in \textit{conv4} and \textit{conv5} layers during initial phase of learning when \textit{fc6}, \textit{fc7} and \textit{fc8} had random initializations. We trained the model with this configuration till $10$ epochs, after which we increased the overall learning rate to $5$x$10^{-3}$. After this, we closely monitored the learning process and controlled learning rates and weight decay manually. We stopped training after $30$ epochs when our validation loss plateaued.

\textbf{Data Agumentation:} In order to prevent model overfitting, we augmented our training dataset by applying random brightness adjustment, contrast adjustment and gaussian noise to each training image. We did not apply cropping because it often removed important semantic cues from images. Similar to \cite{alexnet2012}, we subtracted mean $RGB$ pixel values computed over the entire training dataset from each input image.

\section{Experimental Results and Discussion}
\label{sec:results}
All experiments were performed using Caffe \cite{caffe2014} deep learning framework with NVIDIA GeForce GTX Titan X GPU support. The results of our model are compared with current state-of-the-art method of Ciocca \textit{et al.} \cite{cioccalbp2015}, using the original code provided by the authors. For testing and comparison, we used $58,754$ images from SUN$397$ dataset. For cross-dataset evaluation, we consider MIT Indoor \cite{mitindoor2009}, INRIA Holidays \cite{holidays2008} and Pascal VOC 2012 \cite{pascalvoc2010} datasets. We use recommended test set of $1340$ images from $67$ different indoor scene categories of MIT Indoor dataset. INRIA Holidays dataset originally had $1491$ images; however, we removed several duplicate and orientation ambigous images, leading to final test set size of $1233$ images. Lastly, we used the PascalVOC 2012 training and validation dataset of $6233$ images from $12$ different object categories. In the test datasets, images were initially in correct orientation and for testing purpose, we rotated randomly selected images by $90^{\circ}$, $180^{\circ}$ or $270^{\circ}$. The test datasets were balanced, i.e., each of the four orientation classes had equal number of images in the test datasets.  However, Ciocca \textit{et al.}'s method was evaluated under three conditions: 

\begin{enumerate}
	\setlength\itemsep{0.0em}
    \item{\textit{CC-ORIG}\textemdash All test datasets are created according to the scheme proposed in their paper, $72\%$ images are in $0^\circ$ orientation, $14\%$ in $90^\circ$ and remaining $14\%$ in $270^\circ$.} 

    \item{\textit{CC-BAL}\textemdash All test datasets are balanced, i.e., $34\%$ images are in $0^\circ$ orientation, $33\%$ in $90^\circ$ and rest $33\%$ in $270^\circ$.}

	\item{\textit{CC-OUR} \textemdash We modified their method to include $180^\circ$ orientation and trained as well as tested the method on balanced datasets.}
\end{enumerate}

\begin{table}[t]
	\centering
	{
		\caption{Comparison of accuracy with Ciocca \textit{et al.} \cite{cioccalbp2015}}
		\label{tab:quantcomparison}
		\begin{tabular}{| c | c | c | c | c |}
			\hline
			& \emph{SUN$397$} & \emph{Indoor} & \emph{Holidays} & \emph{Pascal}\\
			\hline
			\emph{CC-ORIG} & 92 & 77.69 & 74.13 & 74.94\\
			\hline
			\emph{CC-BAL} & 64.70 & 55.07 & 39.25 & 42.16\\
			\hline
			\emph{CC-OUR} & 81.69 & 70.52 & 77.29 & 70.64\\
			\hline
			\emph{Ours} & \textbf{95.16} & \textbf{95.28} & \textbf{90.88} & \textbf{90.77}\\
			\hline
		\end{tabular}
	}		
\end{table}

Table \ref{tab:quantcomparison} shows the quantitative results of our model compared to Ciocca \textit{et al.}'s method. The lower accuracies obtained with \textit{CC-ORIG} on MIT Indoor ($77.69\%$), Holidays ($74.13\%$) and PascalVOC ($74.94\%$) datasets show that the method doesn't generalize properly to images outside SUN$397$ dataset. The hand-engineered low-level features do not help the model to generalize properly to images outside the training dataset. Further, when the test datasets are balanced (\textit{CC-BAL}), the accuracy of the method drops drastically even for the SUN$397$ test dataset (only $64.7\%$ from $92\%$). This clearly shows that the method is baised towards correctly oriented images which constitute $72\%$ of training and testing datasets. \textit{CC-OUR} gives average results on all test datasets with accuracy ranging from $70$-$82\%$. The quantitative results of \textit{CC-BAL} and \textit{CC-ORIG} reveal that the $92\%$ accuracy obtained with \textit{CC-ORIG} on SUN$397$ dataset is an artifact of imbalanced training and testing datasets. It also shows that the problem of image orientation detection was considerably simplified in \textit{CC-ORIG} by ignoring $180^\circ$ orientation.

In contrast, our model achieved an impressive accuracy of $95\%$ on SUN$397$ and MIT Indoor datasets, while on INRIA Holidays and PascalVOC $2012$ datasets it achieved approximately $91\%$ accuracy. This shows the remarkable generalization capability of our model which detects correct orientation angle of a large variety of images outside the training dataset. The drop in accuracy in case of Holidays and Pascal VOC $2012$ test datasets can be attributed to broad categories of objects, such as animals, cooking utensils, bicycles, etc., which were either absent or sparse in the training dataset derived from SUN$397$ dataset. The accuracy of $95\%$ on MIT Indoor dataset is quite impressive because existing methods have reported problems with indoor images which contain lots of background clutter and lack discriminative features compared to outdoor images.

Fig.~\ref{fig:qualitativeresults} shows the qualitative results obtained with our model for some of the challenging images from different test datasets. We have also presented visualization of the local image regions \cite{selvarajugradcam2016} which were considered discriminative by our model for the orientation detection task. Images shown in Fig.~\ref{fig:sample4}, \ref{fig:sample6}, \ref{fig:sample13}, \ref{fig:sample14}, \ref{fig:sample15} are quite challenging. In Fig.~\ref{fig:sample13}, the model recognizes ground to correctly orient the confusing image of bamboo trees, in Fig.~\ref{fig:sample14} it identifies occluded person, while in Fig.~\ref{fig:sample15} it discriminates between actual mountains and their reflections. 

The qualitative as well as quantitative results clearly indicate superiority of our CNN model over current state-of-the-art method \cite{cioccalbp2015} which is based on hand-crafted features. It is evident from the evaluation results that the hand-engineered features used in \cite{cioccalbp2015} fail to capture the vast amount of semantic content in images which is required for image orientation detection task. It is also evident from the results that the existing methods considerably simplified the image orientation detection problem. After a rigorous quantitative evaluation of our model on balanced test datasets which also include images in $180^\circ$ orientation, we obtain an impressive average accuracy of $93\%$. This is quite close to human performance ($98\%$), as reported in the psychophysical study \cite{psychophysicalstudy2003}. We found that the performance of our model on noisy, tilted and underwater images was encouraging compared to existing method. Overall, we observed that our model lacked orientation knowledge of objects which were absent or scarce in our training dataset. The performance of our model can be easily improved by extending the training dataset to include different variety of images.

\section{Conclusions and Future Work}
\label{sec:conclusions}
In this work, for the first time, a deep learning based approach for image orientation detection task was proposed. Our fine-tuned convolutional neural network model significantly outperformed the state-of-the-art method in literature. It was shown that the existing methods which mainly reckon on hand-engineered features, fail to generalize properly to images outside the training dataset. It was also shown that the problem of image orientation detection was considerably simplified by existing methods and their performance on real life images is average. In contrast, the proposed model, after extensive evaluation, achieved an impressive maximum accuracy of $95\%$ and average accuracy of $93\%$ which is best till date and is very close to human performance reported in literature. The quantitative as well as qualitative results show the impressive generalizing capability of the proposed deep learning based model for the challenging image orientation detection task. It is shown that unlike existing methods which reckon on hand-engineered features, the performance of the proposed model on real life images is superior and far better. In future, we will work towards enhancing our training dataset, consider other deep learning architectures and work on estimating the exact skew angle of images using a hierarchical approach.

\section{Acknowledgments}
We are thankful to the anonymous reviewers, Dr. Narasinga Rao Miniskar, Dr. Pratibha Moogi and Mr. Anurag Mithalal Jain for their invaluable feedback and suggestions.

\bibliographystyle{IEEEbib}
\bibliography{final_paper}

\end{document}